# Joint Embedding with Dynamic Integration network (JEDI-net)

Rishav Mukherjee[1*]
Jeffrey Ahearn Thompson[1]

[1] Department of Biostatistics & Data Science, University of Kansas Medical Center, Kansas City, Kansas, USA

* Corresponding author
E-mail : rmukherjee@kumc.edu

# Abstract

Deep learning has become the dominant form of machine learning due, in part, to its ability to automatically engineer complex meta features from data. However, even with deep learning, such features can perform erratically due to the presence of unobserved latent factors. Incorporating more explicit latent feature structure into deep learning can improve model performance by allowing the model to alter its behavior based on these inferred latent features While traditional approaches rely on clustering the original feature space, neural networks often capture such complexity inherently, limiting the benefit of external clusteringClustering observations by their feature importance profile, rather than raw features, offers a principled way to uncover such latent structures as observations grouped by similar importance patterns may share underlying predictive dynamics.In a prior work, Feature Oriented Representation with Clustering and Explanation (FORCE), we have established that this approach leads to measurable improvement in neural network performance.

In this article, we present JEDI-net (Joint Embedding with Dynamic Integration Network), a unified neural architecture that periodically computes SHAP values during optimization to identify latent subgroups based on similar patterns of feature importance and embeds these subgroups identities directly into the network, allowing predictions to adapt based on an observation's underlying predictive structure. This eliminates the need for separate explainers or attention components, allowing the model to learn and integrate feature importance in a more efficient and interpretable way. We evaluate JEDI-net on three benchmark datasets and show that it consistently matches or surpasses the performance of FORCE, while reducing complexity. By embedding explanation directly into the learning loop, JEDI-net advances a more integrated and scalable approach to deep model interpretation and prediction.

# Background

Latent structures in data refer to hidden patterns or relationships within a dataset that are not directly observable but can be inferred from the observed data. Latent structures are crucial for understanding the underlying mechanisms of that generate the observed data and are often studied using methods like clustering and prediction. [1, 2] Capturing such latent unobserved structures to improve the accuracy of predictive models has become an active avenue of deep learning research. Most approaches cluster original features in order to infer latent structures under the assumption that observations with similar feature profiles belong to the same underlying subgroup and may share predictive relationships to the outcome[3-5]. However, owing to the ability of deep learning approaches to automatically engineer meta features, [6], not much can be gained from adding on information from clustering observations in the original feature space. In a previous work, we designed a hybrid method, FORCE (Feature Oriented Representation with Clustering and Explanation) that leverages a flexible clustering mechanism in the space of feature importance values rather than the raw features, in an effort to understand if we could derive information about latent structures in the data and use that to improve neural network performance[7]. The idea behind FORCE is that converting the features into importance scores that are related to the outcome by design, would allow the model to learn and gain insights about the underlying subgroups that may have varying predictive relationships to the outcome of interest. FORCE demonstrated that this approach leads to an improvement in performance of deep learning models applied to a range of different data.

The importance metric of choice used in FORCE was SHapley Additive exPlanations (SHAP). SHAP is a method in explainable AI that leverages Shapley values from cooperative game theory to fairly allocate contributions of individual features to model predictions. SHAP values attribute the change in a model's output to individual features, by considering the contribution of each feature across all possible "coalitions" of features; this ensures a fair distribution of feature importance, making it a powerful tool for explaining complex models[8, 9]. In order to derive the SHAP values we chose to fit each of the datasets using a Gradient Boosting Classifier in lieu of its ability to identify patterns in a wide variety of datasets without making assumptions about the distribution of the data and the ability to account for complex interactions[10]. The absolute SHAP values obtained from this method were used in the network architecture for FORCE in two ways viz. cluster labels derived from clustering observations based on their SHAP profiles and a SHAP-initialted attention layer. These two mechanisms have demonstrated capabilities to improve not just the predictive ability of neural networks but have also led to improved robustness in the overall predictive process.

While FORCE was able to establish the idea that SHAP values can be used to identify outcome related latent structures in the data and using that information does lead to an improvement of overall model performance, there was a considerable amount of computational complexity involved in building an "ensemble" method that used a separate predictive process (i.e the GBC) to retrieve the SHAP values, cluster observations using kernel k-means and then initiate a separate attention mechanism within the network architecture. Thus, in this project we set out to develop an integrated model that derives SHAP values internally and uses them to cluster observations by feature importance, without the obverhead of FORCE.

These insights led us to build an architecture, Joint Embedding with Dynamic Integration network (JEDI-net), an integrated framework that uses SHAP values produced by the neural network itself at various stages of training to help the model learn about the latent structures present in the data. In JEDI we eliminated the need for utilizing a separate predictive model before the initiation of the network in order to derive the SHAP values and also a separate attention mechanism. Instead, we incorporate the k-means clustering algorithm within the network architecture. This allows the model to dynamically group observations using their absolute SHAP profiles during model training itself and learn from the evolving process. The use of k-means as a clustering algorithm was made to integrate the information from SHAP values so that they are more accurate and the weights are useful and also to control the complexity of the overall network while still having capabilities to define and group observations based on similarities of the differential impact of features on predictions. JEDI-net uses absolute SHAP values to capture the magnitude of feature influence and encodes cluster assignments as dense embeddings to allow the network to learn and refine meaningful representations of latent subgroups throughout training.

# Methods

A schematic representation of the JEDI-net architecture is shown in Fig 1.

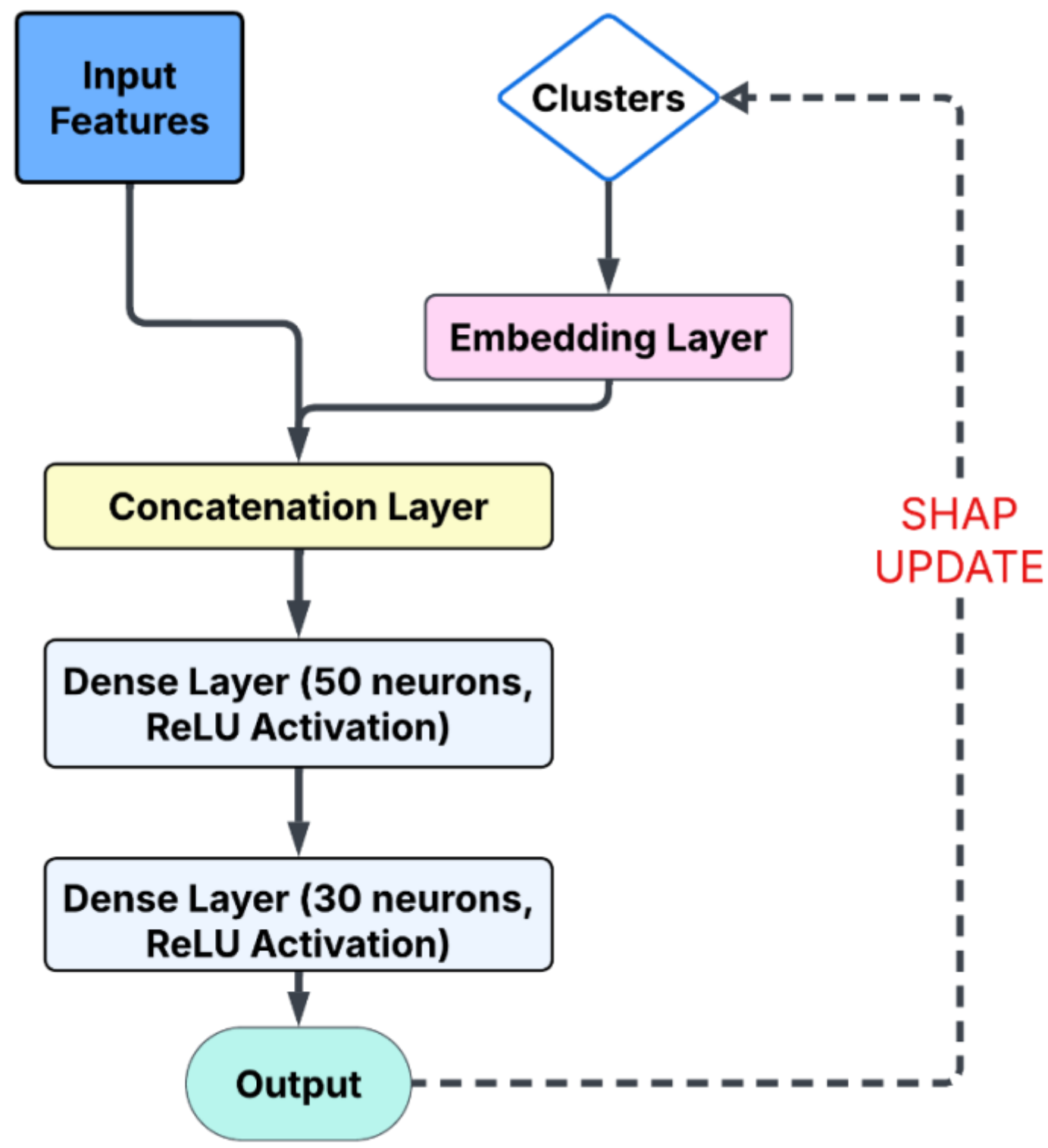

**Fig 1:Schematic of the JEDI-net (Joint Embedding with Dynamic Integration) network**

JEDI-net is a neural network architecture that incorporates cluster-level information dynamically using SHAP values. These clusters represent latent classes which are not observable from the original feature set. The latent classes group observations based on similarity in which features are predictive for them, allowing the model to adapt to this information.

The following steps describe the architecture and forward pass procedure of JEDI-net, from initial data input through dynamic cluster intergration.

*Step 1 : Input data and initial clustering of observations based on original features.*

The network starts off with accepting input features $X \in R^d$ as input. Prior to training the observations are grouped into clusters using KMeans clustering based on their raw input features. These are the initial cluster assignments and are meant to act as placeholders until the first subsequent SHAP update while also contributing training the network on some initial information about the structural similarity in the data.

*Step 2 : Embedding Cluster Labels*

The cluster assignments are encoded using an embedding layer of dimension 5. The embedding layer performs entity based embedding which converts the integer cluster label into a dense vectorial representation of dimension 5. The embedding layer allows the network to learn meaningful representation of clusters that are updated using backpropagation which adjusts the embedding so that clusters with similar predictive behavior have closer representations. The embedding vector encodes each cluster in a way that captures the underlying latent structure within groups, while also learning the relationships between the latent classes themselves.

*Step 3 : Concatenation of cluster embedding with the features and pass through dense layers*

The resulting cluster embedding from Step 2 is concatenated with the input feature vector, allowing the network to incorporate latent subgroup characteristics. The concatenated representation is passed through two dense layers each with 50 and 30 units respectively, followed by a ReLU activation.

*Step 4 : Output for binary classification and SHAP updates*

After passing through the successive dense layers the final output layer consists of a sigmoid activation function for binary classification. To dynamically refine the cluster definitions, we use a SHAP update mechanism. At fixed intervals in training the absolute SHAP values are computed for each observation to quantify feature contributions.While SHAP values can theoretically be computed at every epoch, doing so is computationally prohibitive given the cost of computing SHAP values across all observation; fixed interval updates therefore represent a pragmatic trade-off between cluster refinement frequency and computational feasibility. The use of the absolute SHAP values instead of the raw SHAP values was motivated by the idea that we are interested in the variable's influence on a prediction irrespective of its directionality. The observations are then re-clustered based on their SHAP vectors using KMeans and cluster memberships are updated. The updated cluster labels are realigned with the previous cluster labels using the Hungarian algorithm on the Euclidean distance between the previous and current cluster labels to preserve continuity across training iterations. These labels are then re-encoded using the embedding layer and follows the rest of the forward pass as defined in steps 1, 2 and 3.

## Hungarian Algorithm for cluster assignment

The Hungarian algorithm is a combinatorial optimization technique that is used to solve minimum cost bipartite matching problems in polynomial time[11]. The algorithm creates a cost matrix where the rows represent old clusters, and the columns represent the new clusters with elements of the matrix representing the Euclidean distance between the old and new cluster centroids. Given, this cost matrix , the Hungarian algorithm find the optimal one-to-one correspondence between old and new clusters such that the total Euclidean distance across all matched pairs is minimized. This is done using a combinatorial optimization method that ensures the overall distance is minimized. This results in a one-to-one mapping between the old and new clusters that minimizes disruption from the periodic SHAP updates.

## Dataset Description

Three datasets were used to train and evaluate the performance of FORCE. These were the (i) Early Stage Diabetes Risk Prediction with 520 observations and 16 features, for which the output variable represents the presence or absence of diabetes[12] (ii) The Polycystic Ovarian Syndrome (PCOS) dataset with 1000 observations and 5 features, the output variable is the presence or absence of polycystic ovary syndrome; and (iii) the credit approval dataset which has 690 observations with 14 features and a status variable indicating whether the observations credit card application was approved or not[13]. All datasets were obtained from the UCI Machine Learning repository.

All datasets underwent minimal preprocessing involving standard scaling of continuous features and one hot encoding of categorical features.

## 5-fold Cross Validation to identify cluster cardinality (K) and SHAP update interval.

JEDI-net was trained on each dataset using a nested cross-validation framework, consisting of a 5-fold outer CV and a 3-fold inner CV. The inner loop was used to identify the optimal combination of key hyperparameters, including the number of clusters K (values: 2, 3, 4), the SHAP update interval (every 10, 20, 30, or 40 epochs), and the dimension of the cluster embedding (2, 3, 4, 5). The model was trained for 100 epochs during each evaluation. At each inner iteration, performance was assessed using the F1 score, and the best-performing hyperparameter combination was selected to train the model on the full outer training fold. The final evaluation was performed on the corresponding outer test fold, and metrics including F1 score, accuracy, and area under the ROC curve (AUC) were recorded. This nested procedure ensured a robust estimate of generalization performance while simultaneously tuning hyperparameters in a data-driven manner. Performance was compared against a baseline neural network (Base NN) that did not include SHAP-based dynamic cluster embedding as well as FORCE.

# Results

The performance metrics for each method on all three datasets are shown in Fig. 2. For each of the datasets the modelwas evaluated using the nested cross-validation framework described in Methods section, with hyperparameters selected via the inner loop and final performance recorded on the outer test folds. . The results for the Base NN configuration and FORCE are also presented for comparison. Tables 1-3 present the numerical results of JEDI-net compared to FORCE.

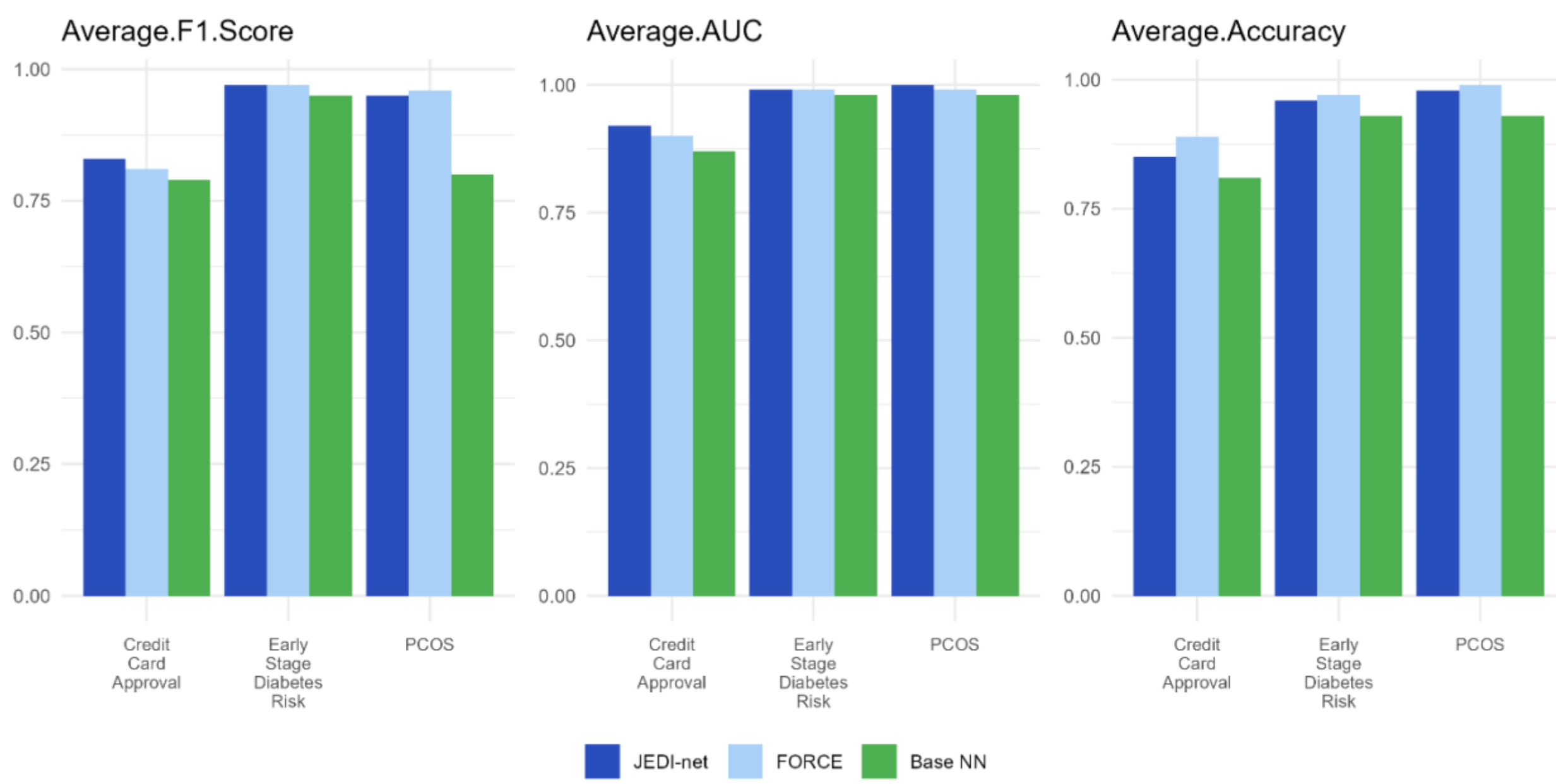

**Figure 2 : Evaluation Metrics for comparing the JEDI-net with FORCE to the model without the SHAP based dynamic cluster embedding (Base NN)**

**Table 1: Evaluation metrics of JEDI-net and FORCE for the Early Stage Diabetes risk detection data.**

| Metric | JEDI-net | FORCE |
|---|---|---|
| F1-Score | 0.97 | 0.97 |
| Accuracy | 0.96 | 0.97 |
| AUC | 0.99 | 0.99 |

JEDI-net performed as good as FORCE in predicting the presence of early stage diabetes.

**Table 2: Evaluation metrics of JEDI-net and FORCE for the PCOS data.**

| Metric | JEDI-net | FORCE |
|---|---|---|
| F1-Score | 0.95 | 0.96 |
| Accuracy | 0.98 | 0.99 |
| AUC | 1.00 | 0.99 |

*JEDI-net performed as good as FORCE in predicting the presence of PCOS.*

**Table 3: Evaluation metrics of JEDI-net and FORCE for the Credit Approval data.**

| Metric | JEDI-net | FORCE |
|---|---|---|
| F1-Score | 0.83 | 0.81 |
| Accuracy | 0.85 | 0.89 |
| AUC | 0.92 | 0.9 |

JEDI-net outperformed FORCE in terms of the F1 score and AUC.

# Discussion

Joint Embedding with Dynamic Integration network (JEDI-net) was developed to address two core limitations of FORCE architecture: its reliance on an external explainer model and the computational burden of kernel-based clustering. While FORCE successfully demonstrated that clustering absolute SHAP values can uncover latent structures related to predictions that improve predictive performance [7], its ensemble/ multi-stage pipeline required a gradient boosting classifier for SHAP generation, kernel k-means for clustering, and a separate attention mechanism. These attributes of the architecture posed opportunities to make the model more streamlined and scalable for real world applications for its utility owing to the self contained architecture.

JEDI-net integrates these stages into a single end-to-end trainable architecture. It uses internal SHAP computations and dynamic clustering updates that evolve alongside the training process. This idea aligns with recent work in explainable AI emphasizing that explanations should not be restricted to be post hoc findings but can also be embedded in the model's learning loop. [14, 15] Our results demonstrate that JEDI-net matches or surpasses FORCE on the three benchmark datasets. The improvement in the credit approval dataset particularly showcases the ability of JEDI-net's dynamic SHAP-based clustering to adapt to shifting data distributions. Although both models were optimized for the F1 score, the improvement in the AUC-ROC across the datasets on using JEDI-net suggests that the model is more robust across a range of cutoffs for each of the binary predictions. The use of the Hungarian Algorithm to maintain consistency in cluster identities across SHAP updates and allows the network to adapt while preserving semantic continuity, a feature, that to our knowledge has not be explored in explainability-guided models.

Both FORCE and JEDI-net use absolute shap values , capturing the magnitude of feature influence rather than its direction. The choice is deliberate; clustering based on magnitude rather than signed contributions produces more stable and interpretable latent groupings, particularly in setings where the same feature may exert differing effects across different observations.[16] JEDI-net inherits this design principle from FORCE , but extends it by encoding the resulting cluster assignments into a dense embedding space that is continuously refined throughout training, allowing the model to not only group observations by shared importance allowing the model to not only group observations by shared feature importance profiled, but to learn the relationships between the groups dynamically.

The use of k-means clustering in place of kernel k-means was a deliberate tradeoff between computational efficiency and clustering granularity. While kernel k-means may capture more complex boundaries the computational complexity increases in polynomial time; $O(n^2)$ rendering the model with some limitations in terms of scalability to larger datasets.[17] In contrast, standard K-means offers linear scalability and, in our experiments, proved sufficient for capturing the latent structures needed to enhance downstream learning. When combined with embedding layers, these latent cluster assignments allow the model to learn and refine relationships during training, enabling adaptive representation learning without the overhead of more complex clustering methods.

Another potentially important aspect is interpretability. By learning cluster embeddings based on absolute SHAP profiles, JEDI-net can potentially allow practitioners to trace both why an observation was classified and which latent subgroup it belongs to. Interpretability was not the focus of this article and thus it will be addressed in future work. Compared with FORCE, which required manual SHAP analysis and external clustering, JEDI-net supports more fluid learning based on evolving data patterns.

JEDI-net also diverges significantly from FORCE in terms of the removal of the attention mechanism. FORCE uses SHAP-initiated attention layers to further refine feature weighting during training, and this has shown to yield performance gains as compared to random weight initialization However, our results suggests that the dynamic SHAP clustering in JEDI-net may implicitly capture some of this variance in feature relevance, allowing us to omit the attention mechanism without substantial sacrifice in predictive performance. This streamlining simplifies the implementation while preserving interpretability. While FORCE occasionally achieves higher raw accuracy—for instance, in the credit card approval task—accuracy alone can be misleading in imbalanced classification problems. In such contexts, metrics like F1 score and AUC provide a more balanced view of model performance. JEDI-net consistently demonstrates competitive or superior performance on these more robust evaluation metrics, which better reflect real-world predictive reliability.

In moving from a two model ensemble approach (GBC+DNN) to a unified model, JEDI-net enables end to end learning and reduces reliance on preprocessing stages which aligns with the broader goals in explainable AI as mentioned earlier in this section [14]. More broadly, JEDI-net can be viewed as an instance of a growing research direction in which structured knowledge or auxiliary signals are embedded directly into the training process to guide representation learning, often referred to as knowledge-guided or informed machine learning [18,19]. A prominent example is physics-informed learning, where physical laws are incorporated as constraints in loss functions or architectures so that learned representations remain consistent with known domain structure [20]. This idea has been applied successfully in graph neural networks, for instance in physics-informed graph convolutional networks for predicting information cascades in complex networks [21]. JEDI-net shares the central premise of this paradigm, that model learning improves when guided by an informative signal beyond the raw input features, but differs in the source of that signal. Rather than importing external prior knowledge, JEDI-net derives its guidance internally: SHAP-based feature-importance profiles are computed during training and used to form dynamic clusters whose embeddings feed back into the learning process. In this sense, it is a "self-guided" variant of knowledge-guided learning, in which the model's own explanatory signals serve as the constraint. Situating JEDI-net in this context shows that its contribution is not limited to a streamlined version of FORCE. It is an application of a general principle, using internal signals to guide model learning, to the tabular setting, where such approaches remain comparatively under-explored. We anticipate that combining internally generated signals with external domain knowledge, such as clinical or physiological constraints, is a promising direction for future work.

However, there are several limitations worth noting. First, the reliance on SHAP values, while now internally computed, still introduces additional computation during training due to periodic updates. Second, while k-means works well in our benchmarks, its assumptions about cluster shapes may not hold in all domains. Future extensions of JEDI-net would explore more flexible clustering methods to further improve adaptability across varied data landscapes. While our current evaluation focused on three publicly available datasets, including those with varying dimensionality and complexity, future work will aim to benchmark JEDI-net to datasets larger in scale and complexity such as

In summary, JEDI-net offers a more integrated, efficient, and interpretable alternative to FORCE. It retains the core philosophy: that learning from latent structures based on feature importance—rather than raw features—can improve both performance and understanding. But by embedding this insight into the network itself, JEDI-net offers a step toward truly self-explaining models and contributes to the broader movement toward knowledge-guided learning in which models are steered by informative internal or external signals. those present in graph-structured data with spatiotemporal dependencies, to heterogeneous networks, or to networks with specific topologies such as tree-like structures [22] . This, we understand, would require additional considerations for mapping such dependancies while generating SHAP based updates and is a highly valuable direction for future work.

# Conclusion

In this article, we introduce JEDI-net (Joint Embedding with Dynamic Integration Network) — a unified model that advances deep learning by embedding SHAP-based clustering directly into the model training process. JEDI-net builds upon the core insights of FORCE, retaining the benefits of explainability-driven supervision while eliminating the architectural overhead of separate explainers and attention layers. By dynamically learning from latent subgroup structures derived from absolute SHAP values, the model is able to adaptively guide training based on evolving patterns of feature importance. Our evaluation across multiple benchmark datasets demonstrates that JEDI-net consistently matches or exceeds the performance of its predecessor, particularly in complex or heterogeneous data environments. These results provide strong evidence that integrating explainable clustering within the training loop not only enhances model interpretability but also leads to measurable gains in predictive performance**.** JEDI-net thus represents a step forward in building models that are both accurate and inherently explainable

# Data Availability:

All datasets used in the article are publicly available and are obtained from UCI machine learning library.
1. Early Stage Diabetes Risk prediction dataset : https://archive.ics.uci.edu/dataset/529/early+stage+diabetes+risk+prediction+dataset
2. Polycystic Ovary Syndrome data : https://www.kaggle.com/datasets/lucass0s0/polycystic-ovary-syndrome-pcos.
3. Credit approval data : https://archive.ics.uci.edu/dataset/27/credit+approval